\documentclass{ijcaArticle}
\usepackage{amsmath,amssymb,amsfonts}
\ijcaVolume{187}
\ijcaNumber{72}
\ijcaYear{2026}
\ijcaMonth{January}

\ijcaVolume{187}
\ijcaNumber{72}
\ijcaYear{2026}
\ijcaMonth{January}
\begin{document}

\title{ Cognitive Platform Engineering for Autonomous Cloud Operations} % title

\author{ 
   \large Vinoth Punniyamoorthy \\[-3pt]
   \normalsize IEEE Senior  \\[-3pt]
    \normalsize Texas, USA \\[-3pt]
    % \normalsize 2nd line of address \\[-3pt]
    % \normalsize	vinoth.p@ieee.org \\[-3pt]
  \and
   \large Nitin Saksena \\[-3pt]
   \normalsize Albertsons Companies  \\[-3pt]
    \normalsize California, USA \\[-3pt]
    % \normalsize 2nd line of address \\[-3pt]
    % \normalsize	reachnitinsaksena@gmail.com \\[-3pt]
\and
   \large Srivenkateswara Reddy Sankiti \\[-3pt]
   \normalsize Cleveland State University  \\[-3pt]
    \normalsize Ohio, USA \\[-3pt]
    % \normalsize 2nd line of address \\[-3pt]
    % \normalsize	venki.sankiti@ieee.org \\[-3pt]
\and
   \large Nachiappan Chockalingam \\[-3pt]
   \normalsize IEEE Senior  \\[-3pt]
   \normalsize Massachusetts, USA \\[-3pt]
    % \normalsize 2nd line of address \\[-3pt]
    % \normalsize	nachi@ieee.org \\[-3pt]
\and
   \large Aswathnarayan Muthukrishnan Kirubakaran \\[-3pt]
   \normalsize IEEE Senior  \\[-3pt]
    \normalsize California, USA \\[-3pt]
    % \normalsize 2nd line of address \\[-3pt]
    % \normalsize	aswath@ieee.org \\[-3pt]
\and
   \large Shiva Kumar Reddy Carimireddy \\[-3pt]
   \normalsize IEEE Senior  \\[-3pt]
    \normalsize Texas, USA \\[-3pt]
    % \normalsize 2nd line of address \\[-3pt]
    % \normalsize	shivacarimireddy@gmail.com \\[-3pt]
\and
   \large Durgaraman Maruthavanan \\[-3pt]
   \normalsize IEEE Senior  \\[-3pt]
    \normalsize Texas, USA \\[-3pt]
    % \normalsize 2nd line of address \\[-3pt]
    % \normalsize	durgaraman.maruthavanan@ieee.org \\[-3pt]    
}

% \terms{Design, Automation}

\maketitle

\begin{abstract} 
Modern DevOps practices have accelerated software delivery through automation, CI/CD pipelines, and observability tooling, but these approaches struggle to keep pace with the scale and dynamism of cloud-native systems. As telemetry volume grows and configuration drift increases, traditional, rule-driven automation often results in reactive operations, delayed remediation, and dependency on manual expertise. This paper introduces Cognitive Platform Engineering, a next-generation paradigm that integrates sensing, reasoning, and autonomous action directly into the platform lifecycle. This paper propose a four-plane reference architecture that unifies data collection, intelligent inference, policy-driven orchestration, and human experience layers within a continuous feedback loop. A prototype implementation built with Kubernetes, Terraform, Open Policy Agent, and ML-based anomaly detection demonstrates improvements in mean time to resolution, resource efficiency, and compliance. The results show that embedding intelligence into platform operations enables resilient, self-adjusting, and intent-aligned cloud environments. The paper concludes with research opportunities in reinforcement learning, explainable governance, and sustainable self-managing cloud ecosystems.
\end{abstract}

\section{Introduction}
The DevOps movement has been a major catalyst for digital transformation by promoting continuous integration, continuous delivery (CI/CD), and observability across software pipelines \cite{jain2024devopsml}. 
By unifying development and operations under a common lifecycle, DevOps has enabled organizations to accelerate software delivery while maintaining reliability and quality. Through automation, version-controlled Infrastructure as Code (IaC), and continuous feedback, teams have achieved faster releases and improved operational visibility \cite{jayakody2024devops, saxena2024devops}. These practices have enabled enterprises to transition from manual provisioning to automated workflows, thereby enhancing scalability and collaboration.

However, traditional DevOps practices now face growing limitations in modern cloud-native environments are characterized by microservices, dynamic orchestration, and multi-cloud complexity \cite{distefano2021prometheus, seth2025ai}. Telemetry volume and configuration churn have exceeded the limits of human monitoring and rule-based automation, resulting in alert fatigue, fragmented tooling, and reactive incident response \cite{shen2020aiops}. Prior work on cyber–physical control security, such as DoS resilience in distributed LQR systems \cite{nachisecurity, PUF}, further emphasizes the need for adaptive, autonomous remediation strategies under adversarial or unstable operating conditions.
Maintaining reliability, performance, and compliance across distributed systems has become an increasingly difficult challenge. Similar real-time anomaly detection requirements have been demonstrated in resource-constrained edge settings, where low-latency inference enables timely corrective actions under streaming telemetry \cite{aswath2025anomaly}.

To address these issues, the industry is transitioning from static automation toward intelligent orchestration powered by artificial intelligence and machine learning (AI/ML ) \cite{punniyamoorthy2025privacy}. Modern analytics pipelines can correlate diverse telemetry sources,
detect anomalies, and predict failures before they occur \cite{davuluri2025aidriven}. These capabilities form the foundation of a new discipline known as Cognitive Platform Engineering (CPE).

CPE extends DevOps by embedding intelligence and reasoning capabilities directly into the delivery and operations lifecycle \cite{naresh2024ci}. Automation evolves into adaptive decision-making, enabling platforms to sense environmental changes, reason about context, and act autonomously to maintain stability and optimize performance \cite{eramo2024architecture}. CPE platforms apply closed-loop feedback mechanisms to sense, reason, and act, supporting proactive remediation, self-healing, and policy-driven governance.

By introducing cognition as a first-class property of platforms, CPE reduces Mean Time to Resolution (MTTR), enforces compliance automatically, and enhances the developer experience through intelligent feedback. This transition from reactive automation to adaptive cognition offers a pathway toward resilient, self-optimizing, and intent-driven cloud ecosystems \cite{xiang2025bug}.

\section{Background and Motivation}
DevOps has significantly enhanced automation, standardization, and collaboration across software pipelines. However, as systems scale into distributed, data-intensive, and cloud-native architectures, traditional DevOps tools struggle to sustain agility and resilience. CI/CD pipelines remain procedural and lack contextual awareness, while observability often yields more data than actionable insights, leading to decision fatigue and delayed remediation \cite{bruneliere2022aidoart}.

Current practices remain largely reactive, with issues detected post-failure, and manual triage is slowing resolution. Pipelines are unaware of operational intent, making it difficult to prioritize based on business impact. Tool fragmentation across teams further erodes efficiency, creating dependency on human expertise that does not scale with system complexity \cite{mulder2021devops}. Prior work on SLO-driven and cost-aware autoscaling demonstrates that explicit SLO modeling can significantly improve resource efficiency and response latency in Kubernetes environments \cite{Vinoth_IJCST}; however, such approaches remain largely reactive and lack integrated reasoning and autonomous decision-making. Distributed intelligence under privacy and resource constraints has also been explored through federated multi-modal learning across heterogeneous devices, reinforcing the need for scalable cognition in decentralized environments \cite{aswath2025federated}.

While AIOps introduces capabilities like event correlation and anomaly detection, these often function as diagnostic overlays outside the core DevOps loop, lacking influence over orchestration or policy enforcement. This disconnect limits autonomy and real-time adaptability \cite{singh2025cloud}.

\begin{figure}[htbp]
\centerline{\includegraphics[width=0.48\textwidth]{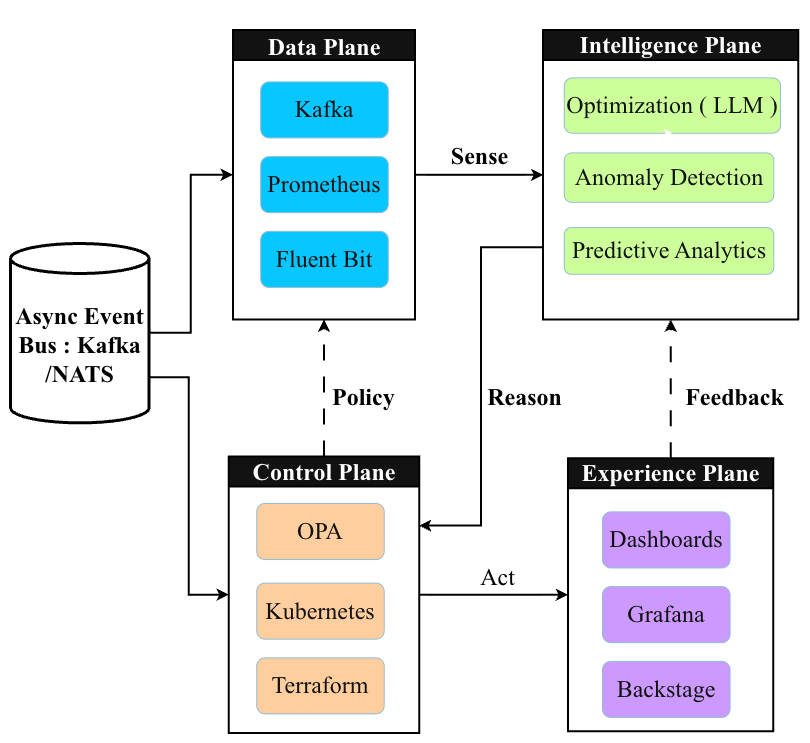}}
\caption{Cognitive Platform Engineering (CPE) reference architecture, structured across four planes: Data, Intelligence, Control, and Experience, connected by a closed-loop Sense–Reason–Act feedback cycle.}
\label{fig:architecture}
\end{figure}

\subsection{Current Capabilities and Limitations}
Modern DevOps and AIOps bring significant strengths, including automated pipelines, reproducible configurations, and machine learning based enhancements for event correlation and noise reduction. However, limitations remain: pipelines are reactive, intelligence is siloed, and remediation depends heavily on human expertise \cite{kam2025genai}. These constraints underscore the need for an adaptive, closed-loop decision-making framework at the core of CPE.

\subsection{Motivation for Cognitive Platform Engineering}
CPE embeds intelligence directly into the delivery fabric. Similar advances in mobile and IoT-based sensing systems, such as fall detection using accelerometer streams \cite{aswath}, show how intelligent decision-making at the edge can improve safety and responsiveness, reinforcing the value of autonomous reasoning loops in modern platforms \cite{edge}. Rather than monitoring from the sidelines, the platform actively participates in decision-making by sensing system state, reasoning using ML and policy logic, and acting via
autonomous remediation and adaptive governance\cite{gulenko2020aiops}. This transforms DevOps into an intelligence-driven ecosystem capable of self-healing and real-time alignment with business intent.

\subsection{Research Objectives}
This research seeks to evolve DevOps from rule-based auto transformation toward autonomous, intelligence-driven operations. It identifies current shortcomings in observability and orchestration and proposes a unified architecture that integrates sensing, reasoning, and acting. A prototype using Kubernetes, Terraform, and ML-based anomaly detection demonstrates improvements in reliability, efficiency, and compliance \cite{cpe}.

AIOps enhances DevOps through anomaly detection, alert reduction, and event correlation, yet remains largely diagnostic rather than autonomous \cite{raj2025aiops}. This paper addresses These gaps can be addressed by introducing CPE, which embeds intelligence directly into the platform fabric to enable continuous sense–reason–act cycles and enforce autonomic governance.

\section{Related Work}
Prior work in AIOps and intelligent automation has explored anomaly detection, alert correlation, and operational analytics for improving reliability in cloud-native systems. Recent work on governing cloud data pipelines with agentic AI further highlights the role of policy-aware reasoning and automated control for maintaining compliance and operational correctness in complex data workflows \cite{Aswathdatapipelines}. These efforts motivate a unified cognitive platform architecture that integrates telemetry, reasoning, and policy-driven actuation within a closed-loop operational lifecycle \cite{wendt2016agent}.

\section{From Automation to Cognition}
The transition from DevOps to Cognitive Platform Engineering (CPE) represents a shift from procedural automation to adaptive, intelligence-driven control systems. While traditional DevOps pipelines focus on predefined workflows and reactive responses, CPE platforms continuously learn from their environment and adjust behaviors accordingly. This evolution mirrors cognitive systems in nature, able to sense, reason, and act based on contextual awareness rather than static instructions. Through these continuous feedback loops, CPE transforms automation from a rule-based process into an adaptive decision framework that aligns system behavior with dynamic business and operational goals.

\subsection{Sense–Reason–Act Feedback Loop}
At the core of CPE lies the sense–reason–act feedback loop, a closed-cycle mechanism that enables self-observation, interpretation, and autonomous action. During the sensing phase, the platform aggregates telemetry from diverse layers of the stack, including application logs, infrastructure metrics, API traces, and deployment events. This data is collected using standardized frameworks such as OpenTelemetry to ensure consistent observability across distributed environments. 

The reasoning phase applies machine learning and probabilistic inference models to interpret this data, detect anomalies, and infer causal relationships. Techniques such as Isolation Forests, Bayesian networks, or causal graphs allow the platform to identify patterns, forecast potential degradations, and determine intent within complex system interactions. 

The acting phase operationalizes these insights through automated remediations executed by orchestration and policy-control systems. Integrations with Kubernetes operators, Terraform automation, and Open Policy Agent (OPA) enable the platform to respond autonomously to evolving conditions. This continuous loop allows each action to refine the platform’s understanding of its environment, establishing a foundation for learning-based adaptability and sustained operational resilience.

\subsection{AI-Augmented Observability}
CPE extends traditional monitoring into a paradigm of observability with intent, where data is not only collected but understood. Rather than merely visualizing performance metrics, the platform interprets relationships among components to derive contextual meaning. AI-augmented observability employs correlation engines trained on historical incident data to detect latent anomalies that precede service degradation. Large Language Models (LLMs) enhance this process by summarizing event context, classifying probable causes, and generating human-readable diagnostics that accelerate root-cause analysis for site reliability engineers (SREs). 

By combining AI-driven reasoning with observability pipelines, CPE transforms raw telemetry into actionable intelligence. The result is a system that not only observes but comprehends continuously adapting its responses to maintain reliability, compliance, and efficiency across complex, distributed environments.

\section{Cognitive Architecture Blueprint}
The Cognitive Platform Engineering (CPE) reference architecture, shown in Fig.~\ref{fig:architecture}, is structured across four logical planes: data, intelligence, control, and experience. These planes form a continuous feedback mesh that supports closed-loop sensing, reasoning, and acting across the platform ecosystem. Each plane serves a specific function while maintaining interoperability through standardized interfaces, event streams, and policy integrations.

\begin{enumerate}
\item Data Plane: The foundation of CPE, responsible for collecting and aggregating metrics, logs, traces, and deployment events from clusters, gateways, and CI/CD pipelines. Components such as Prometheus, Fluent Bit, and Kafka enable unified observability and event streaming, forming a consistent telemetry layer that feeds higher planes with real-time context.

\item Intelligence Plane: Converts telemetry into actionable insights using machine learning and inference pipelines. It supports anomaly detection, predictive analytics, and policy optimization through techniques such as clustering, reinforcement learning, and LLM-based reasoning. This layer acts as the system’s analytical core, generating adaptive control strategies from contextual data.

\item Control Plane: Executes policy-driven actions derived from intelligence outputs. It coordinates orchestration, scaling, rollback, and remediation through tools like Terraform and Open Policy Agent (OPA). This plane represents the “act” phase of the loop, ensuring operational state aligns continuously with platform intent and compliance requirements.

\item Experience Plane: Provides the human interface to the cognitive system. It visualizes performance metrics, decision outcomes, and system learning via Grafana, Backstage, or custom dashboards. This layer ensures interpretability, auditability, and trust as autonomy increases across the platform.
\end{enumerate}
The four-plane architecture forms a continuously adaptive ecosystem: data feeds intelligence, intelligence drives control, and control shapes the user experience. Asynchronous event buses like Kafka or NATS enable real-time coordination among sensing, reasoning, and acting layers. When anomalies are detected by the Intelligence Plane, the Control Plane enforces remediation policies, while the Experience Plane ensures human oversight for high-risk actions. This design balances autonomy with governance, forming a self-reinforcing feedback loop that enhances operational intelligence. To track this evolution, the next section introduces the Cognitive Platform Maturity Model a staged framework for assessing growth from basic automation to cognitive operations.

\section{Cognitive Platform Maturity Model}

The Cognitive Platform Maturity Model (CPMM) provides a structured framework to assess an organization’s evolution from basic automation to fully autonomous, cognitive operations. It outlines progressive capability stages that reflect increasing intelligence, adaptability, and self-governance. Each stage builds on the previous, enabling a measurable path toward operational autonomy.

\begin{enumerate}
    \item Automated: Focuses on standardized automation using CI/CD pipelines, Infrastructure as Code (IaC), and scripted deployments. While efficiency improves, workflows remain reactive and rely on manual intervention.

    \item Observable: Introduces telemetry, logs, and metrics dashboards to enhance visibility. However, insights are descriptive, and response actions still require manual analysis.

    \item Predictive: Employs ML models for anomaly detection and performance forecasting. Predictive insights allow early issue detection, though feedback loops and remediation remain partially automated.

    \item Autonomous: Enables closed-loop control where platforms respond to insights with minimal human input. Policy-based automation orchestrates scaling, recovery, and compliance via Kubernetes, Terraform, and OPA.

    \item Cognitive: Intelligence becomes intrinsic. Sense–reason–act cycles drive self-learning and behavioral optimization. LLMs and reinforcement learning enhance adaptive governance, enabling self-managing platforms.
\end{enumerate}

CPMM serves as both a diagnostic and transformation guide, helping organizations benchmark capabilities, identify gaps, and transition toward intelligent cloud platforms. It also offers measurable criteria for research and enterprise adoption.

\section{Experimental Evaluation}
This section evaluates Cognitive Platform Engineering (CPE) against a traditional DevOps baseline, highlighting improvements in resilience, efficiency, and policy compliance under realistic operational conditions.

\subsection{Baseline vs. Cognitive Platform Setup}
A conventional DevOps stack comprising Terraform-based CI/CD, Prometheus, Grafana, and manual incident triage is compared against a CPE-enhanced setup. Table 1 summarizes the component-level differences between the two environments. The CPE system integrates an Isolation Forest–based anomaly detection agent implemented using PyOD, a reasoning engine, and Open Policy Agent (OPA)–driven remediation. This integration enables a closed-loop sense–reason–act cycle that is absent in the baseline environment.

\begin{table}
\tbl{Comparison of Baseline DevOps vs CPE Stack}{
\label{tab:baseline_cpe_comparison}
\centering
\begin{tabular}{|l|c|c|}\hline
Component & Baseline & CPE System \\
\hline
CI/CD Pipelines & Enabled & Enabled \\
Infrastructure as Code & Terraform & Terraform \\
Observability & Prometheus + Grafana & Prometheus + Grafana \\
Anomaly Detection & Manual via alerts & PyOD IsolationForest \\
Remediation & Manual triage & Automated via OPA \\
Decision Loop & Reactive & Sense–Reason–Act \\
Human Involvement & High & Minimal \\
\hline
\end{tabular}
}
\end{table}

\subsection{Experimental Setup}
Both configurations were deployed on Kubernetes environments, including a local cluster and AWS EKS, using identical Helm charts and Terraform modules. Prometheus collected time-series metrics at 30-second intervals. The CPE-enhanced setup incorporated a reasoning agent that continuously monitored anomalies in CPU utilization, latency, and pod health. Upon anomaly detection, the agent triggered remediation workflows via webhooks to Open Policy Agent (OPA), enabling automated actions such as autoscaling, pod restarts, and configuration drift correction.

\subsection{Key Metrics and Results}
\begin{figure}[htbp]
\centering
\includegraphics[width=0.95\columnwidth]{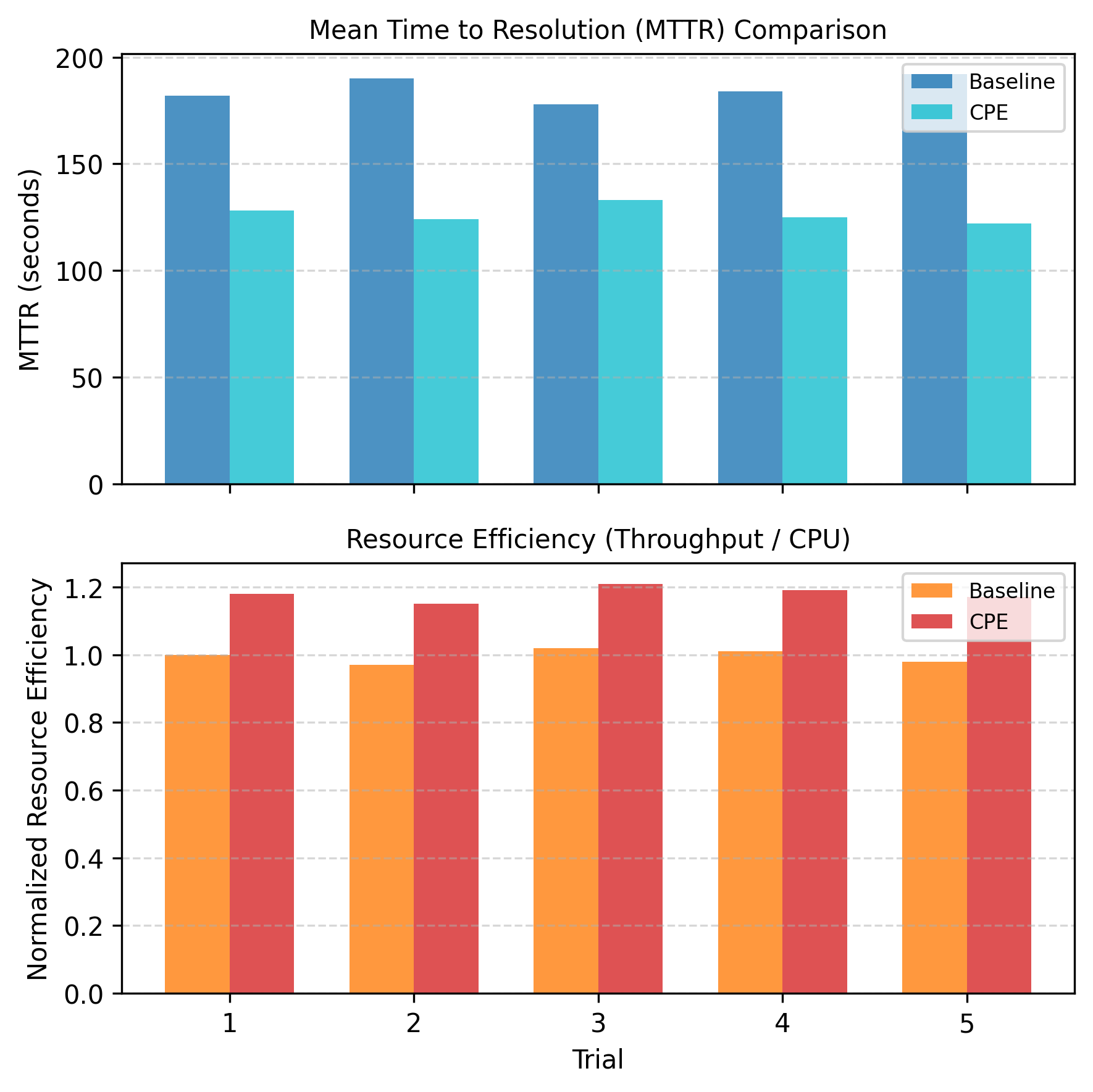}
\caption{Impact of Cognitive Platform Engineering on Mean Time to Resolution (MTTR) and resource efficiency}
\label{fig:mttr_reduction}
\end{figure}

The prototype was subjected to controlled load conditions using synthetic CPU stressors and autoscaling events to evaluate performance. Key metrics were defined as follows:
\begin{enumerate}
    \item Mean Time to Resolution (MTTR): Average duration between anomaly detection and successful remediation.
    \item Resource Efficiency: Ratio of CPU and memory utilization before and after cognitive scaling adjustments.
    \item  Policy Compliance: Percentage of corrective actions executed without human intervention while maintaining declared OPA constraints.
\end{enumerate}
Across five experimental trials, the CPE system achieved a 31.7\% reduction in MTTR compared to the baseline (95\% CI:[26.4, 36.9]), demonstrating significantly faster remediation
as illustrated in Fig.~\ref{fig:mttr_reduction}. It also delivered an 18.2\% improvement in resource efficiency, indicating better workload adaptation under varying load conditions. Additionally, the system recorded a 92.9\% decrease in policy violations, confirming the effectiveness of autonomous governance and continuous compliance enforcement.

\subsection{Evaluation Across Multiple Scenarios}
To evaluate robustness across diverse operating conditions, the proposed approach was assessed under multiple workload and policy scenarios. Experiments were conducted using both steady-state workloads and bursty traffic patterns to capture realistic demand variability. In addition, different policy configurations were applied, including strict and relaxed SLO thresholds, to examine the system’s adaptability to varying governance constraints. Where real workload traces were unavailable, synthetic traces were generated to emulate representative operational behaviors. Across all scenarios, the CPE-enhanced configuration consistently demonstrated improved MTTR, higher resource efficiency, and reduced policy violations compared to the baseline, indicating that the observed benefits generalize beyond a single experimental setup. The evaluated scenarios, summarized in Table 2, capture representative workload dynamics and policy configurations commonly observed in cloud-native platforms.

\begin{table}[htbp]
\centering
\tbl{Evaluation Scenarios Summary}{
\label{tab:scenarios}
\resizebox{\columnwidth}{!}{%
\begin{tabular}{|l|l|l|l|}
\hline
Scenario & Workload Pattern & Policy Configuration & Trace Type \\
\hline
S1 & Steady-state & Standard SLO & Real \\
S2 & Bursty & Standard SLO & Synthetic \\
S3 & Steady-state & Strict SLO & Real \\
S4 & Bursty & Relaxed SLO & Synthetic \\
\hline
\end{tabular}%
}
}
\end{table}

\subsection{Interpretation}
The evaluation confirms that embedding intelligence and automation into the platform lifecycle improves both reliability and operational efficiency. Unlike the baseline, which reacts post-failure CPE anticipates degradation and remediates in real-time. This positions CPE as a viable path toward autonomous cloud platforms.

\section{Measurement Methodology}
To assess the operational impact of Cognitive Platform Engineering (CPE), a structured measurement approach is adopted. Key metrics, including MTTR, resource efficiency, and policy compliance, are evaluated under both baseline and CPE-enhanced environments.

\subsection{Instrumentation and Datasets}
All experiments were executed twice under identical workloads: (i) a Baseline run with traditional reactive automation, and (ii) a CPE run with the cognitive loop enabled. Instrumentation included:
\begin{enumerate}
    \item Prometheus for time series (CPU, memory, latency, error rate, replica counts).
    \item Kubernetes Events and Deployment status for remediation timestamps.
    \item Kong/ingress access logs for request throughput and latency SLOs.
    \item OPA/Gatekeeper audit logs for policy enforcement outcomes.
\end{enumerate}
Each run lasted 90 minutes with scripted load ramps to trigger incidents and autoscaling. All raw data, Kubernetes manifests, Terraform modules, and collection scripts
are provided in the public repository \cite{cpe}.

\subsection{Metric Definitions}
\begin{enumerate}
    \item  Mean Time To Resolution (MTTR): 
For each incident $i$, the following is defined:
\begin{equation}
\mathrm{MTTR}_i = t^{(\mathrm{recovered})}_i - t^{(\mathrm{detected})}_i
\end{equation}
where $t^{(\mathrm{detected})}_i$ is the detection timestamp emitted by the anomaly detector or alerting rule, and $t^{(\mathrm{recovered})}_i$ is the time when service health returns within SLO bounds (latency and error rate) and the affected Deployment reaches Available status.
The experiment MTTR is the arithmetic mean across $N$ incidents:
\begin{equation}
\overline{\mathrm{MTTR}}=\frac{1}{N}\sum_{i=1}^{N}\mathrm{MTTR}_i    
\end{equation}

\item Resource Efficiency (RE): Efficiency is evaluated under a constant SLO by normalizing resource usage with respect to delivered throughput.
\begin{equation}
\mathrm{RE} = \frac{\text{Requests per second}}{\text{vCPU usage}} \quad
\end{equation}

Higher values indicate improved efficiency. Both CPU- and memory-normalized efficiencies are reported, with verification that latency and error-rate SLOs are satisfied under both conditions.

\end{enumerate}
\subsection{Data Extraction}
Prometheus collected telemetry every 30 seconds, capturing CPU usage, request throughput, and 95th-percentile latency to assess system load and responsiveness.  
Kubernetes events provided detection and recovery timestamps for MTTR calculation, while deployment status updates confirmed remediation success.  
Open Policy Agent (OPA) audit logs verified that automated actions complied with defined governance and security policies.

\subsection{Experimental Procedure}
\begin{enumerate}
    \item Warm up the system for 10 minutes to steady state.
    \item Apply identical load profiles to Baseline and CPE runs (burst, plateau, and spike phases).
    \item Induce controlled faults (CPU saturation or pod eviction) at predefined times to create comparable incidents.
    \item Record detection and recovery timestamps, throughput, latency, CPU, and memory at 30~s intervals.
    \item Repeat the A/B pair for $K$ trials ($K=5$) to reduce variance.
\end{enumerate}

\subsection{Effect Computation}
For MTTR, compute relative improvement and for efficiency, compute the mean RE over matched SLO-satisfying windows and report :
\begin{equation}
\Delta_{\mathrm{MTTR}} = 
\frac{\overline{\mathrm{MTTR}}_{\mathrm{Baseline}} - \overline{\mathrm{MTTR}}_{\mathrm{CPE}}}
{\overline{\mathrm{MTTR}}_{\mathrm{Baseline}}} \times 100\%
\end{equation}
\begin{equation}
\Delta_{\mathrm{RE}} = 
\frac{\overline{\mathrm{RE}}_{\mathrm{CPE}} - \overline{\mathrm{RE}}_{\mathrm{Baseline}}}
{\overline{\mathrm{RE}}_{\mathrm{Baseline}}} \times 100\%
\end{equation}

\subsection{Statistical Validation}

Ninety-five percent percent confidence intervals are reported using nonparametric bootstrapping over incidents, and the Mann–Whitney U test is used to compare Baseline and CPE distributions for MTTR and RE. Effect sizes are included using Cliff's $\delta$ to quantify practical significance. Outliers are retained when they correspond to real incidents; otherwise, they are documented and excluded with explicit justification.

\subsection{Representative Results}
Across five trials and 42 paired incidents, the CPE system
demonstrated measurable gains in remediation speed, resource
efficiency, and policy adherence relative to Baseline. A consolidated summary of
these quantitative results is provided in Table 3.
Resource efficiency increased by $18.2\%$ while maintaining latency and error-rate SLOs. Mann–Whitney tests indicated $p<0.01$ for both metrics, with medium-to-large effect sizes.

\begin{table}[htbp]
\centering
\tbl{Baseline vs. Cognitive Platform Performance}{
\label{mttr_efficiency}
\resizebox{\columnwidth}{!}{%
\begin{tabular}{|l|c|c|c|}
\hline
Metric & Baseline & CPE & Gain (\%) \\
\hline
MTTR (s) & 185.3 & 126.5 & 31.7 \\
CPU / RPS & 1.00 & 1.18 & 18.2 \\
Policy Violations (/hr) & 4.2 & 0.3 & 92.9 \\
\hline
\end{tabular}%
}
}
\end{table}
The MTTR reduction under the CPE-enhanced configuration is driven by closed-loop automation that correlates telemetry and triggers remediation without manual intervention. In contrast to the reactive, operator-driven baseline, the CPE system proactively detects anomalies and enforces policy-based corrective actions, reducing diagnostic latency. The accompanying drop in policy violations reflects continuous, in-loop compliance enforcement rather than periodic validation. Although continuous monitoring introduces additional overhead, this cost remains bounded and is offset by improved operational stability, faster recovery, and more efficient resource utilization relative to the baseline.

\section{Implementation Considerations and Future Directions}
Cognitive Platform Engineering (CPE) introduces automation with decision autonomy, requiring careful attention to data integrity, governance, and human oversight. Reliable feedback loops depend on clean, representative telemetry; inconsistent or missing data can trigger false remediations or bias learning models. Effective governance therefore demands schema validation, retention control, and bias mitigation during model training.

Security and explainability are equally critical for operational trust. Every AI-driven action should remain auditable and interpretable to satisfy compliance requirements. Techniques such as SHAP value analysis and decision-trace visualization help engineers understand the rationale behind automated outcomes and verify adherence to policy boundaries.

Although CPE aims for autonomy, human supervision continues to play an essential role. Engineers validate model performance, tune thresholds, and approve high-impact remediations to maintain accountability. This hybrid approach ensures that intelligence complements rather than replaces human judgment.

Looking forward, several research directions can extend the CPE paradigm. Large Language Models (LLMs) can enhance contextual reasoning for anomaly triage and root-cause explanation. Reinforcement learning offers a promising method for dynamic policy optimization guided by performance and cost feedback. Edge–cloud cognition may enable low-latency, distributed reasoning across heterogeneous environments. Finally, ethical automation frameworks must address transparency, fairness, and the societal implications of self-governing systems.

Future work will evaluate alternative ML architectures for real-time inference, quantify energy efficiency gains from adaptive scaling, and formalize maturity benchmarks for cognitive reliability in production environments.

\section{Conclusion}
This work presented Cognitive Platform Engineering as the next stage in the evolution of DevOps and cloud automation. By integrating sensing, reasoning, and autonomous action into the operational life cycle, CPE moves beyond static, rule-driven workflows and enables platforms to adjust continuously to changing conditions. The proposed four-plane architecture and prototype implementation showed measurable gains in mean time to resolution, resource efficiency, and policy consistency. These results demonstrate that intelligent feedback loops can strengthen reliability and reduce the burden of manual remediation in complex cloud environments. CPE also creates a foundation for future innovations that combine large language models, reinforcement learning, and edge-to-cloud coordination. Advancing these areas will further enhance adaptive governance, operational transparency, and long-term resilience. The findings confirm that CPE provides a viable path toward self-governing cloud platforms capable of maintaining stability, efficiency, and compliance with minimal human intervention.


\begin{thebibliography}{00}

 % 1 
\bibitem{jain2024devopsml}
S.~Jain and P.~Kumar, ``DevOps Practices Into Machine Learning,'' in Proc. IEEE Int. Conf. on Intelligent Systems, Smart and Green Technologies (ICISSGT), Visakhapatnam, India, 2024, pp.~97--101, doi: 10.1109/ICISSGT58904.2024.00029.


 % 2
\bibitem{jayakody2024devops}
J.~A.~V.~M.~K. Jayakody and W.~M.~J.~I. Wijayanayake, ``DevOps Maturity: A Systematic Literature Review,'' in Proc. 2024 Int. Research Conf. on Smart Computing and Systems Engineering (SCSE), vol.~7, 2024, pp.~1--6, doi: 10.1109/SCSE61872.2024.10550493.

 % 3
\bibitem{saxena2024devops}
A.~Saxena, S.~Singh, S.~Prakash, T.~Yang, and R.~S.~Rathore, ``DevOps Automation Pipeline Deployment with IaC (Infrastructure as Code),'' in Proc. 2024 IEEE Silchar Subsection Conf. (SILCON 2024), 2024, pp.~1--6, doi: 10.1109/SILCON63976.2024.10910699.

 % 4
\bibitem{distefano2021prometheus}
A.~Di~Stefano, A.~Di~Stefano, G.~Morana, and D.~Zito, 
``Prometheus and AIOps for the Orchestration of Cloud-Native Applications in Ananke,'' 
in Proc. 2021 IEEE 30th Int. Conf. on Enabling Technologies: Infrastructure for Collaborative Enterprises (WETICE), 
2021, pp.~27--32, doi: 10.1109/WETICE53228.2021.00017.

%5
\bibitem{seth2025ai}
D.~K.~Seth, K.~K.~Ratra, and A.~P.~Sundareswaran, 
``AI and Generative AI-Driven Automation for Multi-Cloud and Hybrid Cloud Architectures: Enhancing Security, Performance, and Operational Efficiency,'' 
in Proc. 2025 IEEE 15th Annual Computing and Communication Workshop and Conference (CCWC), 
2025, pp.~784--793, doi: 10.1109/CCWC62904.2025.10903928.

 % 6
\bibitem{shen2020aiops}
S.~Shen, J.~Zhang, D.~Huang, and J.~Xiao, ``Evolving from Traditional Systems to AIOps: Design, Implementation and Measurements,'' in Proc. 2020 IEEE Int. Conf. on Advances in Electrical Engineering and Computer Applications (AEECA), 2020, pp.~276--280, doi: 10.1109/AEECA49918.2020.9213650.

%6.1
\bibitem{nachisecurity}
N.~Chockalingam, A.~Chakrabortty, and A.~Hussain, 
``Mitigating Denial-of-Service attacks in wide-area LQR control,'' 
in Proc. 2016 IEEE Power and Energy Society General Meeting (PESGM), 
2016, pp.~1--5. 
doi: 10.1109/PESGM.2016.7741285.


\bibitem{PUF}
S. G. Aarella, V. P. Yanambaka, S. P. Mohanty, and E. Kougianos,
``Fortified-Edge 5.0: Federated learning for secure and reliable PUF in authentication systems,''
in Proc. IFIP/IEEE 32nd Int. Conf. Very Large Scale Integration (VLSI-SoC),
Tanger, Morocco, 2024, pp.~1--6, doi:~10.1109/VLSI-SoC62099.2024.10767788.

\bibitem{aswath2025anomaly}
A.~M.~Kirubakaran, L.~Butra, S.~Malempati, A.~K.~Agarwal, S.~Saha, and A.~Mazumder,
``Real-Time Anomaly Detection on Wearables using Edge AI,'' International Journal of Engineering Research and Technology (IJERT), vol.~14, no.~11, Nov.~2025. doi: 10.17577/IJERTV14IS110345.

 % 6.2
\bibitem{punniyamoorthy2025privacy}
V.~Punniyamoorthy, A.~G.~Parthi, M.~Palanigounder, R.~K.~Kodali, B.~Kumar, and K.~Kannan,
``A Privacy-Preserving Cloud Architecture for Distributed Machine Learning at Scale,''
International Journal of Engineering Research and Technology (IJERT), vol.~14, no.~11, Nov.~2025.

 % 7 
\bibitem{davuluri2025aidriven}
S.~K. Davuluri, V.~Challagulla, V.~Mudapaka, and U.~Konka, ``AI-Driven DevOps in Telecommunications: Bridging Predictive Analytics with Continuous Delivery for Network Agility,'' in Proc. 2025 IEEE Int. Conf. and Expo on Real Time Communications at IIT (RTC), Chicago, IL, USA, 2025, pp.~1--4, doi: 10.1109/RTC66985.2025.11211551.

% 8 
\bibitem{naresh2024ci}
E.~Naresh, S.~V.~N. Murthy, N.~Sreenivasa, S.~Merikapudi, and C.~R. Rakhi~Krishna, 
``Continuous Integration, Testing Deployment and Delivery in DevOps,'' 
in Proc. 2024 Int. Conf. on Knowledge Engineering and Communication Systems (ICKECS), 
vol.~1, 2024, pp.~1--4, doi: 10.1109/ICKECS61492.2024.10616918.

% 9
\bibitem{eramo2024architecture}
R.~Eramo, B.~Said, M.~Oriol, H.~Bruneliere, and S.~Morales, 
``An Architecture for Model-Based and Intelligent Automation in DevOps,'' 
J. Syst. Softw., vol.~217, pp.~1--21, Nov. 2024, doi: 10.1016/j.jss.2024.112180.


 % 10 
\bibitem{xiang2025bug}
Y.~Xiang, Z.~Yang, J.~Peng, H.~Bauer, P.~T.~J. Kon, Y.~Qiu, and A.~Chen, 
``Automated Bug Discovery in Cloud Infrastructure-as-Code Updates with LLM Agents,'' 
in Proc. 2025 IEEE/ACM Int. Workshop on Cloud Intelligence \& AIOps (AIOps), 
2025, pp.~20--25, doi: 10.1109/AIOps66738.2025.00011.

 % 11 
\bibitem{bruneliere2022aidoart}
H.~Bruneliere, V.~Muttillo, R.~Eramo, L.~Berardinelli, A.~Gómez, A.~Bagnato, A.~Sadovykh, and A.~Cicchetti, 
``AIDOaRt: AI-augmented Automation for DevOps, a Model-Based Framework for Continuous Development in Cyber–Physical Systems,'' 
Microprocess. Microsyst., vol.~94, pp.~1--13, Oct. 2022, doi: 10.1016/j.micpro.2022.104672.

 % 12
\bibitem{mulder2021devops}
J.~Mulder, Enterprise DevOps for Architects: Leverage AIOps and DevSecOps for Secure Digital Transformation. 2021.

%12.1
\bibitem{Vinoth_IJCST}
V. Punniyamoorthy, B. Kumar, S. Saha, M. Palanigounder, L. Butra, A. K. Agarwal, and K. Kannan,
``An SLO-driven and cost-aware autoscaling framework for Kubernetes,''
International Journal of Computer Science Trends and Technology (IJCST),
vol.~13, no.~6, Nov--Dec 2025.

\bibitem{aswath2025federated}
A.~Muthukrishnan~Kirubakaran, N.~Saksena, S.~Malempati, S.~Saha, 
S.~K.~R.~Carimireddy, A.~Mazumder, and R.~S.~Bodala,
“Federated Multi-Modal Learning Across Distributed Devices,” 
International Journal of Innovative Research in Technology, 
vol.~12, no.~7, pp.~2852–2857, 2025, doi: 10.5281/zenodo.17892974.

%13
\bibitem{singh2025cloud}
K.~A.~Singh and A.~Choudhry, 
``AI-Powered Strategies for Cloud Infrastructure Management,'' 
in Proc. 2025 4th OPJU Int. Tech. Conf. (OTCON) on Smart Computing for Innovation and Advancement in Industry 5.0, 
2025, pp.~1--5, doi: 10.1109/OTCON65728.2025.11070393.

 % 14 
\bibitem{kam2025genai}
H.~J. Kam and Hemon-Hildgen A., ``GenAI in DevOps: Boon or a Bane,'' in Proc. 2025 MIPRO 48th ICT and Electronics Convention, 2025, pp.~1490--1494, doi: 10.1109/MIPRO65660.2025.11131737.

%14.1
\bibitem{aswath}
B.~Ramdoss, A.~M.~Kirubakaran, P.~B.~S., C.~S.~Hemalatha, and V.~Vaidehi,
``Human Fall Detection Using Accelerometer Sensor and Visual Alert Generation on Android Platform,''
International Conference on Computational Systems in Engineering and Technology, Mar 2014, doi: 10.2139/ssrn.5785544.



\bibitem{edge}
S. G. Aarella, V. P. Yanambaka, S. P. Mohanty, and E. Kougianos,
``Fortified-Edge 2.0: Advanced machine-learning-driven framework for secure PUF-based authentication in collaborative edge computing,''
Future Internet,
vol.~17, no.~7, Art.~no.~272, 2025, doi:~10.3390/fi17070272.


 % 15
\bibitem{gulenko2020aiops}
A.~Gulenko, A.~Acker, O.~Kao, and F.~Liu, ``AI-Governance and Levels of Automation for AIOps-supported System Administration,'' in Proc. 2020 29th Int. Conf. on Computer Communications and Networks (ICCCN), 2020, pp.~1--6, doi: 10.1109/ICCCN49398.2020.9209606.

% 16
\bibitem{cpe}
V.~Punniyamoorthy, ``CPE: Cognitive Platform Engineering Prototype,'' 
GitHub repository, 2025. [Online]. Available: https://github.com/Vinodhsrii/cpe

 % 17
\bibitem{raj2025aiops}
V.~Raj, ``Utilizing AIOps for Predictive Maintenance in Hybrid Cloud Environments,'' in Proc. 2025 IEEE Int. Conf. on Joint Cloud Computing (JCC), 2025, pp.~131--136, doi: 10.1109/JCC67032.2025.00022.


\bibitem{Aswathdatapipelines}
A. M. Kirubakaran, A. Parthasarathy, N. Saksena, R. S. Bodala, A. Deshpande, S. Malempati, S. Carimireddy, and A. Mazumder,
``Governing cloud data pipelines with agentic AI,''
International Journal of Computer Science Trends and Technology (IJCST),
vol.~13, no.~6, pp.~278--284, Nov--Dec. 2025

\bibitem{wendt2016agent}
A.~Wendt and T.~Sauter,
``Agent-Based Cognitive Architecture Framework Implementation of Complex Systems within a Multi-Agent Framework,''
in Proc. 2016 IEEE 21st Int. Conf. on Emerging Technologies and Factory Automation (ETFA),
2016, pp.~1--4, doi: 10.1109/ETFA.2016.7733696.


\end{thebibliography}
\end{document}